\documentclass[10pt]{article}
\usepackage{amssymb}
\usepackage{amsmath}
\usepackage{amsfonts}
\usepackage{authblk}
\usepackage{graphicx}
\usepackage{subfigure}
\usepackage{algorithm}
\usepackage{algpseudocode}
\usepackage{indentfirst}
\newcommand{\e}{\mathsf{E}}

\title{
Clearing noisy annotations for computed tomography imaging
}
\author[1]{R.Khudorozhkov}
\author[1]{A.Koryagin}
\author[1]{A.Kozhevin \footnote{E-mail: a.kozhevin@analysiscenter.ru}}
\affil[1]{Data Analysis Center\footnote{https://github.com/analysiscenter}}

\begin{document}

\maketitle
\pagestyle{plain}

\begin{abstract}

One of the problems on the way to successful implementation of neural networks is the quality of annotation. For instance, different annotators can annotate images in a different way and very often their decisions do not match exactly and in extreme cases are even mutually exclusive which results in noisy annotations and, consequently, inaccurate predictions.

To avoid that problem in the task of computed tomography (CT) imaging segmentation we propose a clearing algorithm for annotations\footnote{The algorithm is implemented in RadIO library - https://github.com/analysiscenter/radio.}. It consists of 3 stages: 
\begin{itemize}
\item \textit{annotators scoring}, which assigns a higher confidence level to better annotators;
\item \textit{nodules scoring}, which assigns a higher confidence level to nodules confirmed by good annotators;
\item \textit{nodules merging}, which aggregates annotations according to nodules confidence.
\end{itemize}

In general, the algorithm can be applied to many different tasks (namely, binary and multi-class semantic segmentation, and also with trivial adjustments to classification and regression) where there are several annotators labeling each image.

\end{abstract}

\section{Introduction}

With exponential data growth it becomes possible to construct high-performance systems based on neural networks. Unfortunately, one cannot simply collect a lot of data and feed it into neural network. Supervised methods of machine learning require labeled data which are scarce and not infrequently are far from being perfect.

To make annotation more accurate, a dataset can be labeled by several annotators but that is where new problems arise. For instance, different annotators can annotate the same image in different ways and very often their decisions are not similar and sometimes even mutually exclusive. 

For some tasks the methods were proposed, e.g. for image classification one can use algorithm described in \cite{Whitehill}. We concentrate on the task of segmentation. In general case, it's not obvious how to understand that the annotators label the same object on the image or that the object is incorrectly labeled.
In order to perform merging of the same objects and to clear annotation of objects that was founded by few annotators we develop a clearing algorithm for CT images. The algorithm takes into account some specifics of CT images and masks but the main idea can be applied to the wide range of tasks.

\section{Algorithm}

The CT image annotation consists of the information about cancer nodules each of which is represented as a ball. Thus the annotations looks as a table with the columns
image ID, doctor (annotator) ID, Z coordinate of the nodule (in mm), Y coordinate, X coordinate and nodule radius (in mm). Moreover, there is an information about doctors who annotate each image. Generally speaking, that information doesn't contain in annotation table because the doctor could see an image but decided that the patient doesn't have cancer nodules. However, one can put it into the table by adding row with image ID, doctor ID and NaN for all other columns. In the perfect case all annotators who label the same image find the same objects but in the real world their decisions can be very different so the task is to clear very noisy annotation: remove wrong nodules and merge nodules that are the same.

The filtering algorithm consists of the following steps:
\begin{itemize}
\item \textit{annotators scoring}. 
Each annotator gets his personal score in the interval $[0, 1]$. The scoring algorithm is based on the following hypothesis: perfect annotators make the same labels, whereas the worst ones tend to disagree with the others and their annotation always differ. The score can be interpreted as the confidence in decision of that annotator. Note that nobody knows the ground truth so we have to compare annotators between each other to recover it. The main idea is to perform multiple consultations when the annotation of the one annotator is compared with the annotations of two others.
\item \textit{objects (nodules) scoring}.
Each annotated object gets its personal score. The score takes into account distance to other objects and scores of the corresponding annotators. This is needed because even good annotators can mark the same nodule slightly differently and we need some kind of averaging.
\item \textit{objects (nodules) merging}.
That step is needed to merge the same object from annotations of different doctors.
\item \textit{removing of noise objects (nodules)}.
All object with final score which is less then some threshold (say, 0.1) are removed from the annotation.
\end{itemize}

The described algorithm is not limited to the CT images. The only one proposition is that annotators perform detection of objects on the images. For example, the algorithm can be applied to satellite images or nucleus detection. The idea of annotators scoring can be implemented even for the classification task when several annotators label objects with classes. Below we describe all the steps required for the segmentation task.

\section{Annotators scoring}

To estimate the quality of annotation we use the notation of doctors panels (or consiliums). The score of the doctor is an averaged similarity metric between his annotation and annotation of two other doctors in the panel. For a binary segmentation 'Dice' makes a good choice as a similarity metric. A simplified version of the algorithm is described as Algorithm \ref{alg_simpl}.

\begin{algorithm}[!ht]
\caption{Simplified scoring algorithm}\label{alg_simpl}
\begin{algorithmic}[1]
\For{each doctor $d$}
\For{each image $i$ that was annotated by the current doctor}
\For{each pair of doctors $(d_1, d_2)$ who annotate image $i$ and doesn't contain doctor $d$}
\State compute $Dice(M_d, \frac{1}{2}M_{d_1} + \frac{1}{2}M_{d_2})$
\EndFor
\EndFor
\State{average all computed Dice-metrics and put it as a score of the doctor $d$}
\EndFor
\end{algorithmic}
\end{algorithm}

In order to decrease the influence of doctors with lower scores we modify the algorithm and describe iterative procedure as Algorithm \ref{alg_iter}. Let $s^k_d$ be a score of the $d$-th doctor after $k$ iterations of the algorithm, $M_d$ be a binary mask of the $d$-th doctor for some image. Note that simplified Algorithm \ref{alg_simpl} is the first step of the iterative procedure.

\begin{algorithm}[!ht]
\caption{Iterative scoring algorithm}\label{alg_iter}
\begin{algorithmic}[1]
\State{put all scores $s^0_d = 0.5$}
\For{k in \{0, 1, 2, \dots, N-1\}}
\For{each doctor $d$}
\For{each image $i$ that was annotated by the current doctor}
\For{each pair of doctors $(d_1, d_2)$ who annotate image $i$ and doesn't contain doctor $d$}
\State compute $Dice\left(M_d, \frac{s_{d_1}^k}{s_{d_1}^k + s_{d_2}^k} M_{d_1} + \frac{s_{d_2}^k}{s_{d_1}^k + s_{d_2}^k} M_{d_2}\right)$
\EndFor
\EndFor
\State{average all computed Dice-metrics and put it as a $s_d^{k+1}$}
\EndFor
\EndFor
\end{algorithmic}
\end{algorithm}

Let the number of annotators be $p \in \mathbb{N}$, the number of images be $m \in \mathbb{N}$. Algorithm \ref{alg_iter} also can de described by formula
\begin{equation}
s^{k+1}_d = \e Dice \left( M, w^k_1 M_1 + w^k_2 M_2 \right), \,\,\,\, d \in \{1, \dots, p\} 
\end{equation} \label{expectation}
where
\begin{itemize}
	\item $Dice(A, B) = \frac{2|A B|}{|A| + |B|}$ for $A = (a_{ij})_{i=1,j=1}^{n,k}$, $B = (b_{ij})_{i=1,j=1}^{n,k}$, $AB = (a_{ij} b_{ij})_{i=1,j=1}^{n,k}$, $|C| = \sum_{ij} c_{ij}$,
    \item $(I_d, D_{d1}, D_{d2})$ is a random vector uniformly distributed on the set $C_d$ of all possible consiliums for $d$-th doctor, where $I_d$ is an index of the image which was labeled by the $d$-th doctor, $D_{d1}, D_{d2}$ are two other doctors who annotate the same image $I_d$
    \item $M^i_k$ is a binary mask for $i$-th image from the $k$-th annotator, $M = M^{I_d}_d$, $M_1 = M^{I_d}_{D_{d1}}$, $M_2 = M^{I_d}_{D_{d2}}$,
    \item $w^k_1 = \frac{s^k_{D_1}}{s^k_{D_1} + s^k_{D_2}}$, $w_2 = \frac{s^k_{D_2}}{s^k_{D_1} + s^k_{D_2}}$.
\end{itemize}

Initially, all confidences are 0.5. It means that the confidence for all annotators is the same and there is no an any information about their quality. The formula \eqref{expectation} can be also represented as

$$
s^{k+1}_d = \frac{1}{S} \sum_{i \in A_d} \sum_{
\substack{d_1, d_2 \in B_{i,d} \\ 
d_1 < d_2}
} Dice \left( M^i_d, \frac{s^k_{d_1}}{s^k_{d_1} + s^k_{d_2}} M^i_{d_1} + \frac{s^k_{d_2}}{s^k_{d_1} + s^k_{d_2}} M^i_{d_2} \right), \,\,\,\, d \in \{1, \dots, p\},
$$
where $A_d$ is the set of images that were annotated by the $d$-th annotator, $B_{i,d}$ is the set of annotators who annotate the image $i$ except the $d$-th annotator and 
$$
S = \frac{1}{S} \sum_{i \in A_d} \sum_{
\substack{d_1, d_2 \in B_{i,d} \\ 
d_1 < d_2}} 1.
$$

\section{Nodules scoring}
During this stage of the algorithm we score each object from the annotation. Assume that the scores of annotators $(s^1, \dots, s^p)$ are already computed. In order to estimate confidences of nodules $\left\{C(n_j)\right\}_{j=1,\dots, J}$, we use the approach reminiscent of kernel density estimation \cite{kde}. Let 
\begin{itemize}
\item $K(u)$ be a kernel function with a finite support, e.g. Epanechnikov kernel \cite{Ep},
\item $D(n)$ be the annotator who annotated an object $n$ and $s_{D(n)}$ be the score of this annotator, 
\item $\mathcal{S}(n)$ be the scan, to which the object $n$ belongs,
\item $\mathcal{N}(D)$ be the set of objects annotated by an annotator $D$,
\item $r(n, n')$ be the euclidean distance between a pair of objects $n$ and $n'$,
\item $\alpha$ be a coefficient between $0$ and $1$. 
\end{itemize}
Then the confidence $C(n_j)$ of object $n_j$ is given by
\begin{gather}\label{nod_scor}
C(n_j) = \alpha s_{D(n_j)} + (1 - \alpha)\sum\limits_{n \in \mathcal{S}(n_j) \setminus \mathcal{N}(D(n_j)) } K(r(n_j, n)) s_{D(n)}.
\end{gather}
In other words, an object is attributed high confidence if $1)$ it is annotated by a good annotator and 2) in its proximity there are objects, annotated by other trustworthy annotators. Note that algorithm (\ref{nod_scor}) contains parameter $\alpha$. In our experiments we set $\alpha$ to $0.7$, as it ensures good clusterisation of annotated objects in two groups, the ones with acceptably high confidence and the ones that can be deemed erroneous.

\section{Nodules merging}
In the final part of the algorithm we merge different annotations of the same objects. Suppose that an $i$-th image is annotated by doctors $(D_1, \dots, D_n)$. This gives us $n$ different sets of annotated objects $\left(\mathcal{N}(i, D_1), \dots, \mathcal{N}(i, D_n)\right)$. Note that each object has an associated score-value. The purpose of this part is to aggregate $n$ annotations into one annotation $\mathcal{N}(i)$.
This can be done in two steps:
\begin{itemize}
\item Form groups of objects $\left(g_r =\{n_1, \dots, n_{l_r}\}\right)_{r=1 \dots t}$.
\item Merge elements of each $g_r$ into one object $n_r$. Get final annotation $\mathcal{N}(i) = \{n_1, \dots, n_r\}$. Assign each nodule from $\mathcal{N}(i)$ its own confidence.
\end{itemize}
\subsection{Nodules grouping}
We begin with constructing an \textit{overlap graph} $G$. This graph has $N$ vertices, where $N$ is the total number of objects from annotations $\left(\mathcal{N}(i, D_1), \dots, \mathcal{N}(i, D_n)\right)$. In graph $G$, a pair of vertices is connected, whenever the associated nodules overlap. Distributing objects into groups comes down to finding connected components in $G$ (see Figure \ref{merging:a}).
\subsection{Merging elements of groups}
\begin{figure}[!ht]
\begin{center}
\subfigure[Grouping overlapping nodules]{
  	\includegraphics[height=60mm]{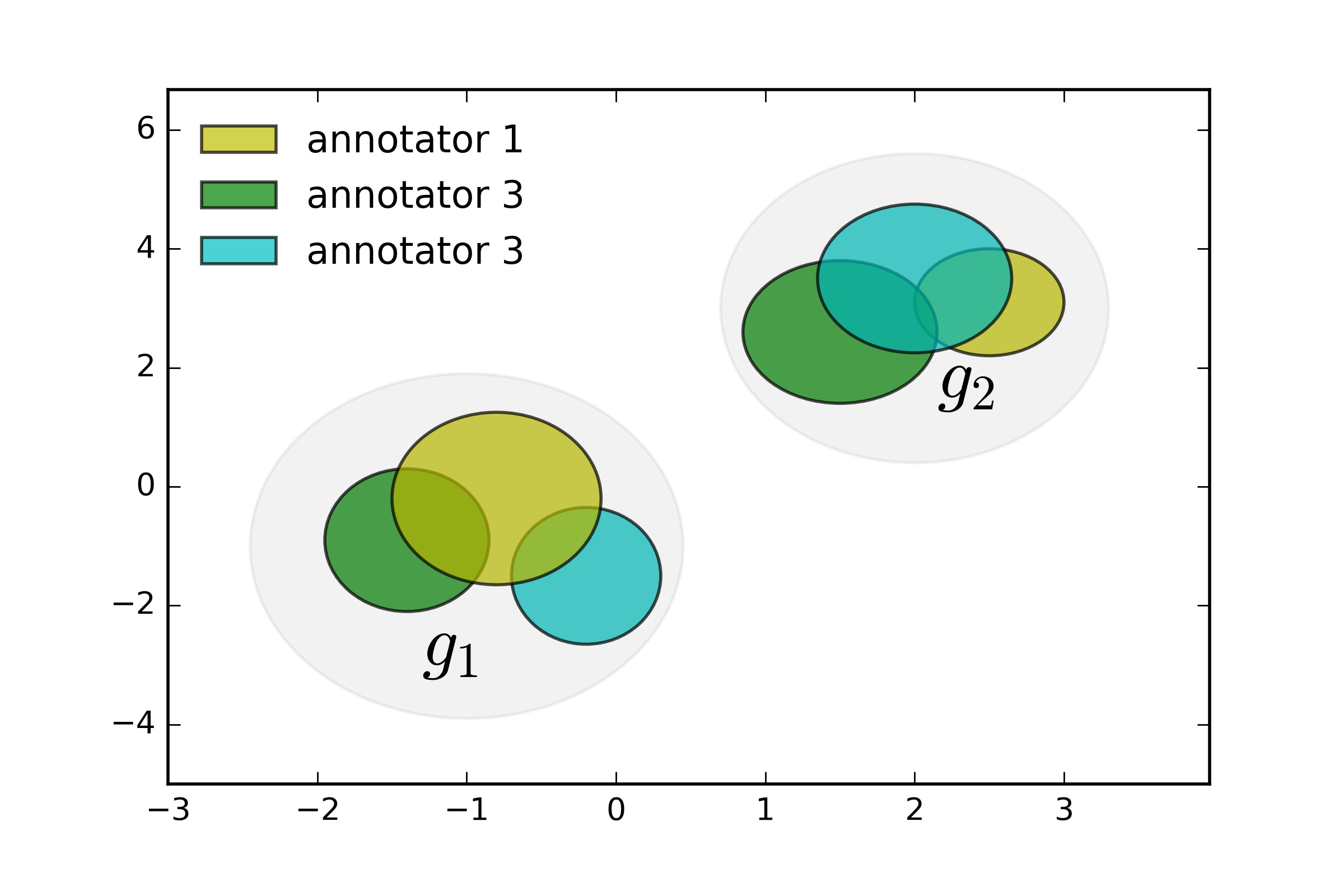}
    \label{merging:a}
  }%
  \subfigure[Merging $g_r$]{
  	\includegraphics[height=60mm]{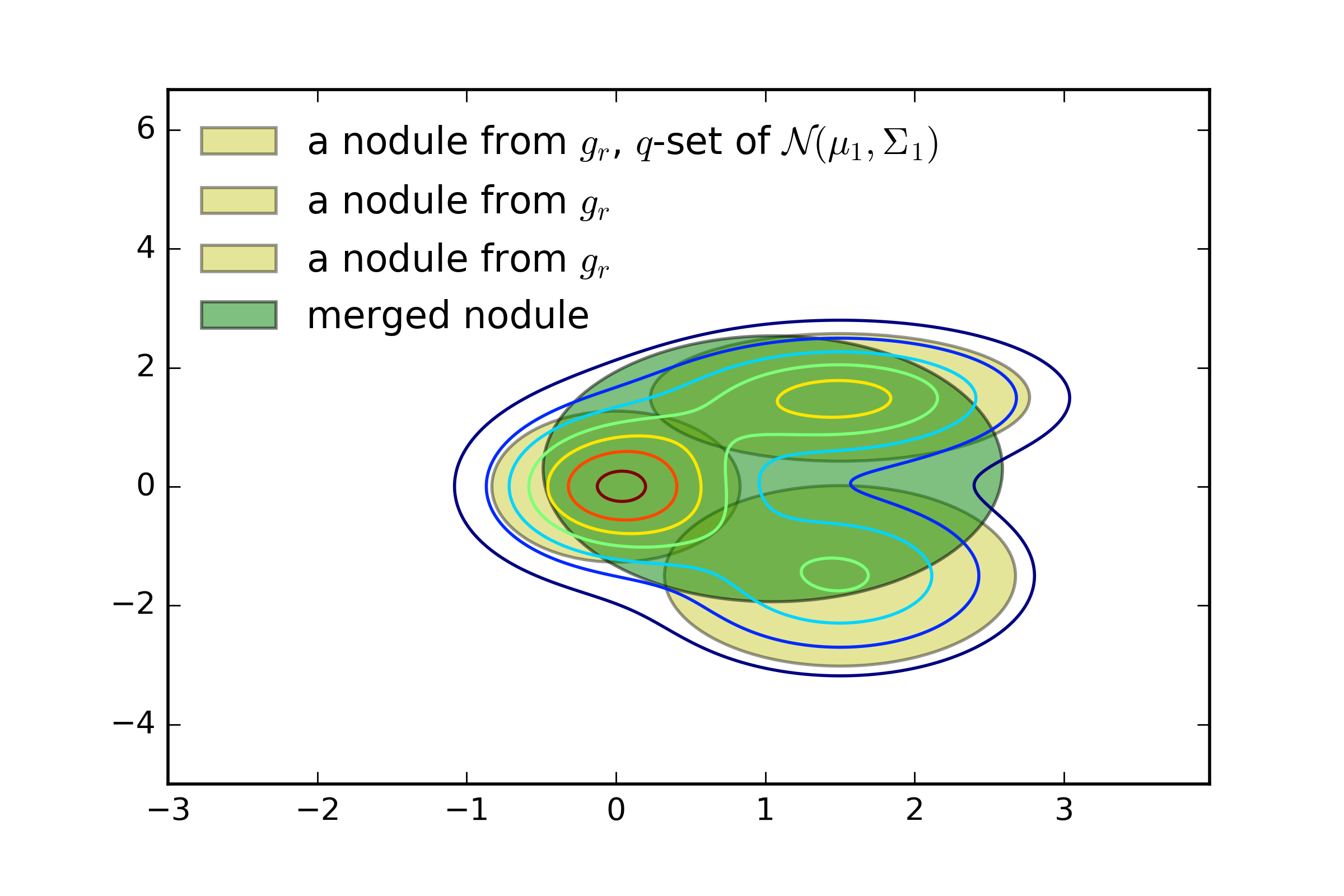}
    \label{merging:b}
  }
  \caption{Merging nodules.}
  \label{merging}
\end{center}
\end{figure}

Firstly, we associate a multivariate normal distribution $d(n)$ with each object $n$ from aggregated annotation. For a nodule with center $(x, y, z)$ and radiuses $(r_x, r_y, r_z)$ we do this in the following manner:
\begin{gather}\label{qset}
d(n) = N(\mu_n, \Sigma_n),\ \mu_n = (x, y, z),\ \Sigma_n=diag(\alpha r_x^2, \alpha r_y^2, \alpha r_z^2); \\
\textrm{Prob}\left[d(n) \in \mathbb{B}((x, y, z); (r_x, r_y, r_z)) \right] = q.
\end{gather}
Note that in the formula above $q$ is a parameter of the algorithm and $\mathbb{B}(xs; rs)$ is an ellipsoid with center $xs$ and radius $rs$. That is, with each nodule we associate a normal distribution, that contains this nodule as a quantile set of probability $q$, with $q$ being the only parameter of the algorithm. In our experiments, we set it to $0.5$. Going further, each group of overlapping nodules $g_r$ can be attributed a mixture of multivariate normal distributions in the following way:
\begin{gather*}
g_r =\{n_1, \dots, n_{l_r}\}_{r=1 \dots t},\\
M(g_r) = \sum\limits_{j=1}^{l_r} w_j d(n_j) = \sum\limits_{j=1}^{l_r} w_j N(\mu_{n_j}, \Sigma_{n_j}),\\
w_j = \frac{C(n_j)}{\sum\limits_{j=1}^{l_r} C(n_j)},\ \ j=1,\dots,l_r.
\end{gather*}
That is, the mixture is composed from gaussians defined earlier, while the weights of separate components are proportional to nodules' confidences $C(n_j)$.
The next step is to aproximate each mixture $M(g_r) = \sum\limits_{j=1}^{l_r} w_j N(\mu_{n_j}, \Sigma_{n_j})$ by gaussian $N(g_r)$ with diagonal $\Sigma$. In doing so, we optimize the KL-divergence between the gaussian and the target mixture:
\begin{gather}\label{optmn}
N(g_r) = \textrm{arg} \min D_{KL}(N \| M(g_r)),\quad N \textrm{ is a gaussian with diagonal } \Sigma.
\end{gather}
In fact, the problem (\ref{optmn}) can be solved analytically. It is straightforward to check that
\begin{gather*}
M(g_r) = \sum\limits_{j=1}^{l_r} w_j N(\mu_{n_j}, \Sigma_{n_j}),\\
N(g_r) = N(\mu_{g_r}, \Sigma_{g_r}),\quad
\mu_{g_r} = \sum\limits_{j=1}^{l_r} w_j \mu_{n_j};\\
\Sigma_{n_j}=\textrm{diag}(\sigma_{n_j}^2),
	\quad \Sigma_{g_r}=\textrm{diag}
    	\left(
			\sum\limits_{j=1}^{l_r} w_j
            	\left[
            		\sigma_{n_j}^2 + (\mu_{n_j} - \mu_{g_r})^2
                \right]
    	\right).
\end{gather*}
The final stage of $g_r$-merging consists of building an artificial nodule as a $q$-quantile set of $N(g_r)$:
\begin{gather*}
N(g_r) = N(\mu_{g_r}, \Sigma_{g_r}) = N(\mu_{g_r}, \sigma_{g_r}^2);\\
n(g_r) = \mathbb{B}\left(\mu_{g_r}, \sqrt{\sigma_{g_r}^2 / \alpha}\right),\\
\alpha \textrm{  s.t.  Prob}\left(N(g_r) \in n(g_r)\right) = q. 
\end{gather*}
Note that this is a reverse of the procedure described in (\ref{qset}). Nodule fitting is demonstrated on the Figure \ref{merging:b}. Finally, the confidence $C(n(g_r))$ is set to the maximum of confidences over all nodules from $g_r$.
\section{Results}
Our algorithm's performance might be demonstrated on LUNA dataset \cite{Luna}. The annotation for the dataset is already cleared so we put it as the ground truth and generate some noisy annotations. To do that we create 3 groups of annotators:
\begin{itemize}
\item BA (\textit{"bad" annotators}) always give wrong annotations,
\item NA (\textit{"normal" annotators}) give both wrong and true nodules,
\item PA (\textit{"perfect" annotators}) give true nodules only.
\end{itemize}

That division is a simplified reflection of reality and allows us to demonstrate the properties of the proposed clearing algorithm.
The generating algorithm described as Algorithm \ref{alg2}.

\begin{algorithm}[!ht]
\caption{Noising algorithm}\label{alg2}
\begin{algorithmic}[1]
\For{each image in dataset}
\For{i in 1, \dots, 10}
\If{i is a "bad" or "normal" annotator}
\State generate $n_i \sim Unif(0, \dots, 4)$
\State generate $n_i$ points $\sim N(\mu, \Sigma)$ and $n_i$ diameters $\sim Unif(4, \dots, 15)$
\State append nodules with the resulting centers and diameters into the new annotation
\EndIf

\If{i is a "normal" or "perfect" annotator}
\For{each nodule in annotation for the current image}
\State add $N(0, \Sigma')$ to the center of the nodule and add $N(0, \sigma')$ to the diameter
\State with probability $p$ add the nodule into the new annotation
\EndFor
\EndIf
\EndFor
\EndFor
\end{algorithmic}
\end{algorithm}
In that algorithm $\mu \in \mathbb{R}^3$ is the center of the current image and
$$
\Sigma =
\left(\begin{matrix}
100 & 0 & 0 \\
0 & 100 & 0 \\
0 & 0 & 200
\end{matrix} \right).
$$
All other parameters vary and depend on the setting.
We consider two settings:
\begin{itemize}
\item (1) \textit{without location-noise:} $\Sigma'= \mathbb{O}_3, \sigma'=0$ and $p = 1$,
\item (2) \textit{with location-noise:} $\Sigma'= \mathbb{I}_3, \sigma'=0.5$ and $p = 0.7$,
\end{itemize}
where $\mathbb{O}_3$ is a zero matrix of size 3 and $\mathbb{I}_3$ is a unit matrix of size 3. The first setting describes situation when all true nodules selected by normal and perfect annotators identically. The second means that true nodules selected by normal and perfect annotators with some Gaussian noise in coordinates and diameter.

We also distribute annotators between groups in different ways:
\begin{itemize}
\item (A) \textit{without middle}: BA = \{0, 1\}, PA = \{2, 3, 4, 5, 6, 7, 8, 9\},
\item (B) \textit{with middle}: BA = \{0, 1\}, NA = \{2, 3\}, PA = \{4, 5, 6, 7, 8, 9\}.
\end{itemize}

This totals to 4 scenarios: A1, A2, B1, B2. And all of them are quite bad, since bad annotators generate a lot of incorrect annotations. And cases A2 and B2 are much worse as  even "perfect" doctors detect nodules not perfectly.

\begin{figure}[!ht]
  \subfigure[A1]{
  	\includegraphics[height=40mm]{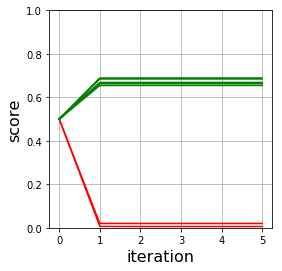}
  }%
  \subfigure[A2]{
  	\includegraphics[height=40mm]{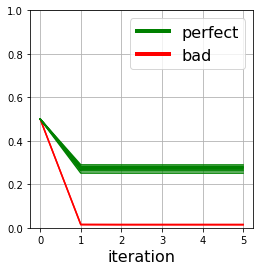}
  }%
  \subfigure[B1]{
  	\includegraphics[height=40mm]{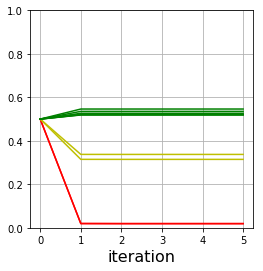}
  }%
  \subfigure[B2]{
  	\includegraphics[height=40mm]{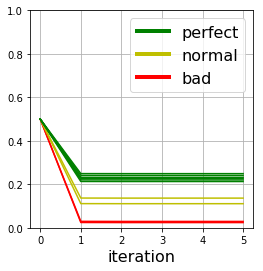}
  }
  \caption{Doctors scoring for different scenarios.}
  \label{fig_scoring}
\end{figure}

Score computations are demonstrated in the Figure \ref{fig_scoring}. As we can see, in all cases algorithm divides annotators into groups which coincide with underlying groups of bad, normal and perfect doctors. That algorithm converges very fast with just 2 iterations to get robust scores.

In cases A1 and B1 the doctors scores are greater since each score is a mathematical expectation of Dice and without location noise Dice is closer to 1. And in cases A2 and B2 prefect doctors are not perfect at all. 

\begin{figure}[!ht]
\begin{center}
\subfigure[]{
  	\includegraphics[height=60mm]{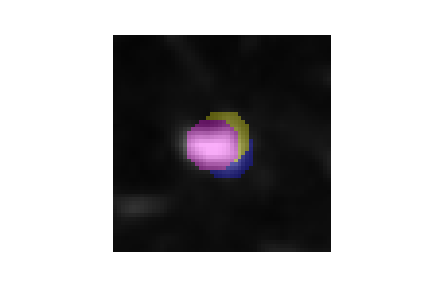}
    \label{fig2:a}
  }%
  \subfigure[]{
  	\includegraphics[height=60mm]{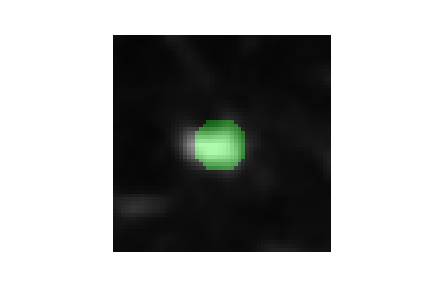}
    \label{fig2:b}
  }
  \caption{Results of nodule merging. \subref{fig2:a}: nodules from 3 annotators,  \subref{fig2:b}: a resulting nodule after merging.}
  \label{fig2}
\end{center}
\end{figure}

Then we compute nodules scores, merge them and remove nodules with final confidence less then 0.1. The example of merging is provided by Figure \ref{fig2}.

Comparison of the resulting annotation and the initial ground-truth provided by Tables \ref{tab1}-\ref{tab3}. We compute pixelwise sensitivity, specificity and intersection-over-union (IoU) of noised and cleared annotation against the original ground-truth annotation. In noised cases we accumulate information from all annotators into one binary mask for each image.

\begin{table}[!ht]
\begin{tabular}{cc}
	\begin{minipage}{.33\linewidth}
    \begin{center}
      \begin{tabular}{| l | l | l |}
      \hline
      Setting & Noised & Cleared \\
      \hline
      A1 & 1.0 & 1.0 \\
      \hline
      A2 & .840 & .800 \\
      \hline
      B1 & 1.0 & 1.0 \\
      \hline
      B2 & .852 & .770 \\
      \hline
      \end{tabular}
      \caption{Sensitivity}
      \label{tab1}
    \end{center}
    \end{minipage} %
    
    \begin{minipage}{.33\linewidth}
    \begin{center}
      \begin{tabular}{| l | l | l |}
      \hline
      Setting & Noised & Cleared \\
      \hline
      A1 & $3.32 \cdot 10^{-6}$ & $3.00 \cdot 10^{-9}$ \\
      \hline
      A2 & $7.93 \cdot 10^{-6}$ & $2.32 \cdot 10^{-6}$ \\
      \hline
      B1 & $6.07 \cdot 10^{-6}$ & $2.82 \cdot 10^{-6}$ \\
      \hline
      B2 & $1.12 \cdot 10^{-5}$ & $2.04 \cdot 10^{-6}$ \\
      \hline
      \end{tabular}
      \caption{1 - Specificity}
      \label{tab2}
    \end{center}
    \end{minipage} %
    
    \begin{minipage}{.33\linewidth}
    \begin{center}
      \begin{tabular}{| l | l | l |}
      \hline
      Setting & Noised & Cleared \\
      \hline
      A1 & .787 & .999 \\
      \hline
      A2 & .504 & .687 \\
      \hline
      B1 & .646 & .799 \\
      \hline
      B2 & .413 & .662 \\
      \hline
      \end{tabular}
      \caption{IoU}
      \label{tab3}
    \end{center}
    \end{minipage} %
\end{tabular}
\end{table}

As we can see, cleared annotations become more similar to the ground-truth annotations: in all settings specificity and IoU is considerably larger. Sensitivity of cleared annotations is controlled by the parameters of the merging procedure.

\section{Summary}
On the one hand, multiple annotators allow us to get more consistent annotation and to avoid mistakes of any single annotator. On the other hand, it is not obvious how to join annotations from several annotators, considering the fact that some annotations are erroneous.

For that purpose we propose an annotation clearing algorithm which scores and merges multiple annotations. It has been successfully implemented for lung cancer datasets where a lot of doctors mark cancer nodules on CT images. In general case, it can be easily modified for other tasks such as multi-class semantic segmentation, object detection, as well as regression and classification.

\end{document}